\documentclass[conference]{IEEEtran}
\usepackage{cvpr}
\usepackage{times}
\usepackage{epsfig}
\usepackage{graphicx}
\usepackage{amsmath}
\usepackage{amssymb}
\usepackage{authblk}
\usepackage{caption}
\usepackage{subcaption}

\usepackage{graphicx}
\usepackage{caption}
\usepackage{subcaption}
\usepackage{amsmath}
\usepackage{setspace}
\usepackage{algorithm}
\usepackage[english]{babel}
\usepackage[utf8]{inputenc}
\usepackage{algorithm}
\usepackage[noend]{algpseudocode}

\usepackage{epstopdf}
\usepackage{amssymb}
\usepackage{authblk}
\usepackage{diagbox}
\usepackage{threeparttable}
\usepackage{graphicx}
\usepackage{booktabs}
\usepackage{chngcntr}
\usepackage{multirow}
\usepackage{cite}
\usepackage{float}
\usepackage{threeparttable}
\usepackage{amssymb}
\usepackage{caption}

\usepackage[breaklinks=true,bookmarks=false]{hyperref}

\cvprfinalcopy 

\def\cvprPaperID{****} 

\begin{document}

\title{Diversity encouraged learning of unsupervised LSTM ensemble for neural activity video prediction}

\author[1]{Yilin Song }
\author[2]{Jonathan Viventi }
\author[1]{Yao Wang }
\affil[1]{Department of Electrical and Computer Engineering,  New York University, NY, USA}
\affil[2]{Department of Biomedical Engineering, Duke University, Durham, NC, USA}


\maketitle

\begin{abstract}

Being able to predict the neural signal in the near future from the current and previous observations has the potential to enable real-time responsive brain stimulation to suppress seizures. We have investigated how to use an auto-encoder model consisting of LSTM cells for such prediction.
 Recognizing that there exist multiple activity pattern clusters, we have further explored to train an ensemble of LSTM models so that each model can specialize in modeling certain neural activities, without explicitly clustering the training data. We train the ensemble using an ensemble-awareness loss, which jointly solves the model assignment problem and the error minimization problem. During training, for each training sequence, only the model that has the lowest reconstruction and prediction error is updated. Intrinsically such a loss function enables each LTSM model to be adapted to a subset of the training sequences that share similar dynamic behavior. We demonstrate this can be trained in an end-to-end manner and achieve significant accuracy in neural activity prediction. 

    \end{abstract}

\section{Introduction}
Epilepsy has been studied for decades and epilepsy surgery outcomes have not improved over 20 years. One third of 60 million people with epilepsy now have seizures that cannot be controlled with medication. Currently existing neurological data analysis rely on manual inspection and most of automatic analysis approaches still depend on clever constructed features like spectral power, wavelet energy spike rate and so on
\cite{bandarabadi2015epileptic,li2013seizure,eftekhar2014ngram,gadhoumi2012discriminating, netoff2009seizure, chua2009automatic,sorensen2010automatic,temko2011eeg,acharya2011automatic}. These methods focus on electroencephalogram (EEG) or electrocorticographic (ECoG) data with coarse spatial and temporal resolution and predict seizure onset relying on several seconds to minutes recording. Recent development of high resolution micro-electrocorticographic ($\mu$ECoG) \cite{delay_map1} unveils rich spatial and temporal patterns. It is tempting to try to predict neural activities in the near feature (milliseconds) to provide guidance for responsive stimulation. Since neural activities are highly non-linear, prediction is quite challenging. To the best of our knowledge we are the first to tackle this problem. 

Recent advances in deep learning provide useful insights in time series prediction. Models for time series prediction and sequence generation could be divided into two major categories: \\
1) models that rely on recurrent neural network (RNN) and its variants \cite{finn2016unsupervised,kalchbrenner2016video,xingjian2015convolutional,srivastava2015unsupervised}. 
2) models that rely on adversarial training \cite{goodfellow2014generative,mathieu2015deep,vondrick2016generating}. 
In deep learning setup, a model is trained end-to-end with appropriate loss function. Supervised learning has been extremely successful in learning good representations \cite{karpathy2015visualizing,zeiler2014visualizing} that could be transferred to other dataset. However videos have much longer duration and detailed annotation down to short time horizon is difficult if not impossible. Researchers have been exploring to characterize the spatial temporal informations in video in an unsupervised manner. Like the pioneering unsupervised LSTM encoder-decoder framework proposed in \cite{srivastava2015unsupervised}, most of the RNN based approaches have an encoder learning a compact feature representation from an input sequence with a decoder reconstructing the input sequence and a predictor predicting future using feature. Other variants of RNN based approaches have modifications on the computation units. In \cite{xingjian2015convolutional}, the authors propose a convolutional LSTM module to better model spatial relationships. In \cite{finn2016unsupervised}, the constructed multiplicative units eliminate the distinction between memory and hidden states in LSTM. Models extended from adversarial training \cite{goodfellow2014generative} have a generative model and discriminative model. The generative and discriminative model are trained in a combatting manner, with the discriminative model predicting whether one instance (frame) is generated by the generative model or comes from dataset. In \cite{mathieu2015deep}, to deal with the blurry predictions resulting from minimizing the mean square error, the authors propose a different loss function and demonstrated adversarial training could be successfully employed for next frame prediction.  

\begin{figure}
        \centering
       	\begin{subfigure}[b]{0.5\textwidth}
                \includegraphics[trim={1cm 2cm 1cm 3cm},clip,width=\textwidth]{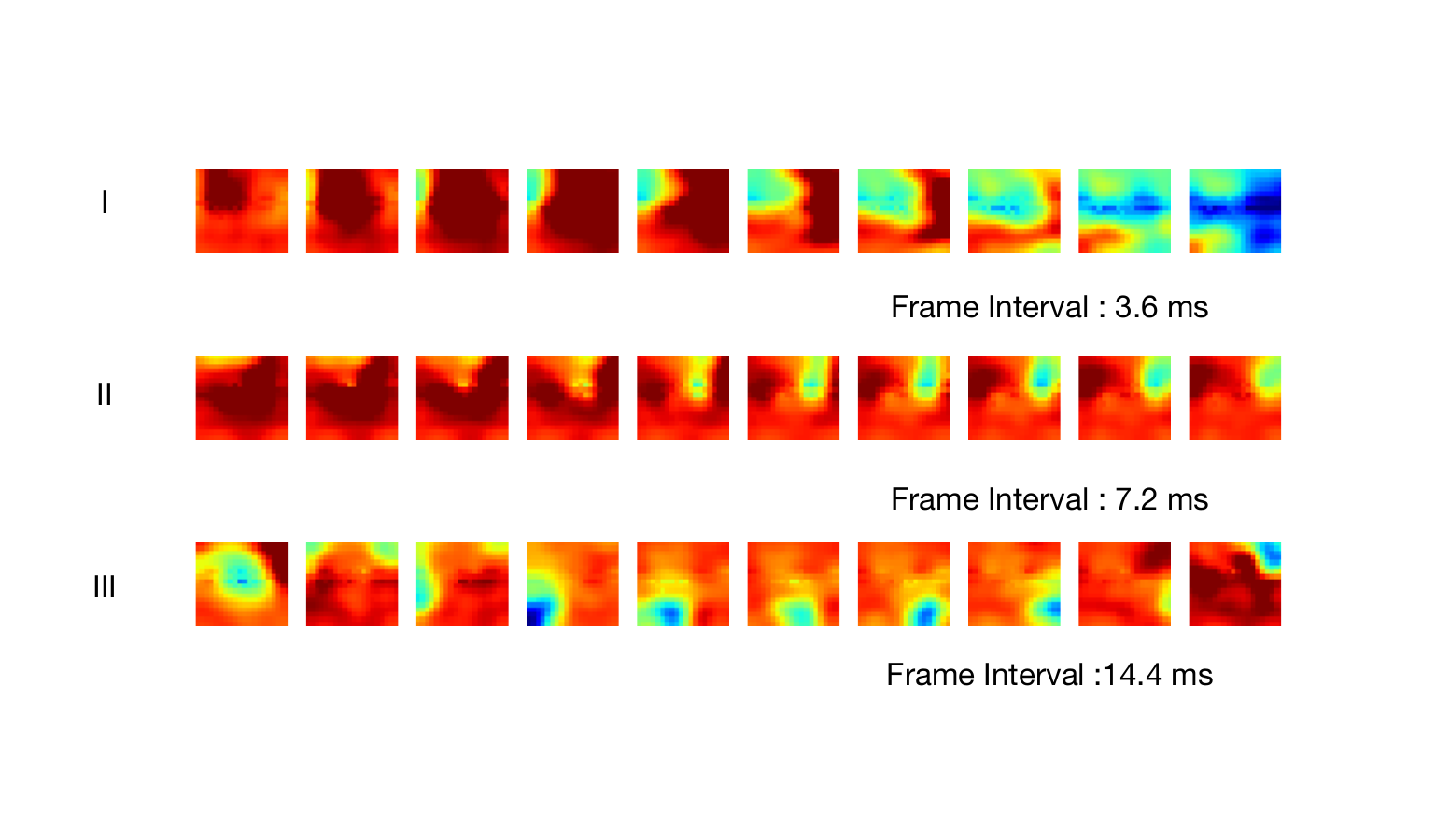}
                \caption{}
        \end{subfigure} 
       	\begin{subfigure}[b]{0.5\textwidth}
                \includegraphics[trim={1cm 2cm 1cm 1cm},clip,width=\textwidth]{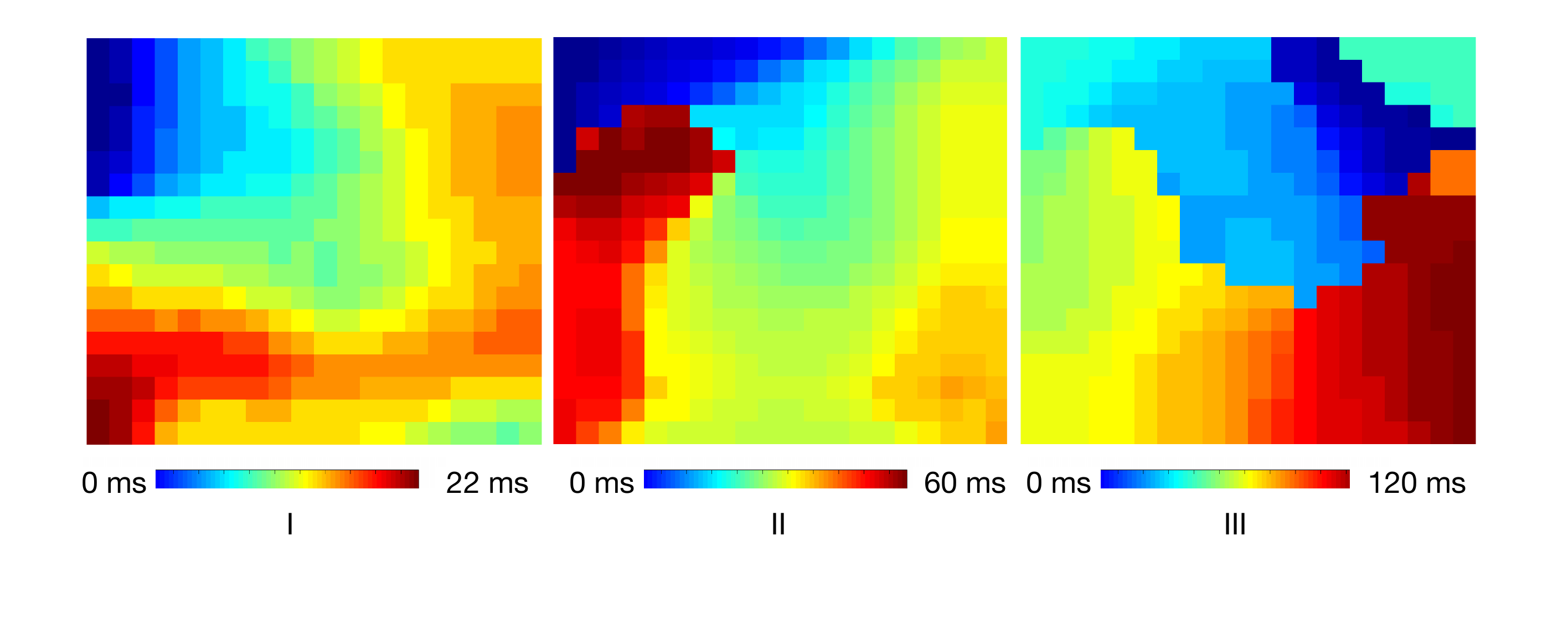}
                 \caption{}
        \end{subfigure}
         \caption{(a) Movie frames shows spatial-temporal patterns from all 360 electrode (18 by 20 grid), the frame interval are selected to fit 10 frames. (b) Delay maps for the corresponding sequences. We illustrate the activity pattern of a video using is delay map following \cite{delay_map1}. Using the average of all channel signals as a reference signal, the delay of each channel with respect to the reference is calculated, and the color at each pixel in a delay map indicates the delay of the corresponding channel relative to the reference signal. Delay map images have the exact same dimensionality as the grid layout. For example, in sequence III, the delay map reveals the counter-clockwise motion of the blue region.}
\label{sequence and delay}
\end{figure}

The neural activities harvested with $\mu$ECoG (Fig.~\ref{sequence and delay}) share some common traits as natural videos, yet unlike natural videos the patterns of neural activities in a local brain region are restricted by neuron connectivity. Such restrictions lead to a finite number of typical patterns such as plane waves and spiral waves as observed in \cite{delay_map1}.
To exploit the multi-cluster nature of such neural activities, we use multiple choice learning (MCL) \cite{lee2015m,guzman2012multiple} to predict neural activities and let each model certain patterns without explicitly clustering the data. Unlike most ensemble models that enhance performance by averaging independently trained models with random initializations, our ensemble is trained using an ensemble-awareness loss function \cite{lee2015m}, which jointly solve the assignment problem and minimization problem. During training, for each given sample sequence, we calculate the reconstruction error and prediction error using each model, and update only the model that has the lowest reconstruction and prediction error. This updating rule encourages diversities between the trained models. Intrinsically preforms clustering while minimizing the ensemble loss. We demonstrate an ensemble of LSTMs can be trained simultaneously using such loss function in an end-to-end manner and achieve significant higher accuracy in neural activity prediction compare to a single LSTM with similar total number of parameters. In \cite{guzman2012multiple}, the authors showed the image classification accuracy gain through MCL training of a set of CNN models, yet during testing time selecting exactly which CNN classification model to use is difficult. However in video prediction setup, with the decoder reconstructing the input sequence, we can determine the reconstruction error using each model, and choose the model that yields the least reconstruction error to perform prediction.  We show that this model selection criterion could achieve comparable prediction accuracy compared to ''oracle" selection. We also develop a separate classifier that decides which model to use for prediction based on encode features all models. We found that this classifier decision further improve the prediction accuracy.  

\section{Methodology}
In this section, we describe the baseline LSTM model we use and the formulation of multiple choice learning. 
\subsection{LSTM}
For general purpose of sequence modeling, LSTM as a special RNN structure has proven to be capable of modeling long range dependency in many applications \cite{graves2013generating,xu2015show,sutskever2014sequence}. The crucial feature of LSTM compared to the classical RNN to RNN is the memory cell noted by $c_t$, which serves as a conveyor belt connecting time series and acts as an accumulator. The input gate $i_t$ controls the extent that current input $x_t$ and past hidden states $h_{t-1}$ have on affecting the current cell state. Simultaneously a sigmoid layer called forget gate $f_t$ decides what information is going to be thrown away or dampened from the current cell state. Finally the output of the LSTM is a filtered version of the current cell state controlled by the output gate $o_t$ and pushed through a tanh function so that the output has values between -1 and 1. The basic LSTM cell structures are summarized as follows, where $\circ$ denotes the Hadamard product:
\begin {equation}
\begin{aligned}
i_t &= \sigma( W_{xi} x_t + W_{hi} h_{t-1}+W_{ci}\circ c_{t-1}+b_i) \\
f_t &= \sigma(W_{xf}x_t +W_{hf}h_{t-1} + W_{cf}\circ c_{t-1} +b_f)\\
c_t &= f_t\circ c_{t-1}+i_t\circ \tanh(W_{xc}x_t+W_{hc}h_{t-1}+b_c)\\
o_t &= \sigma(W_{xo}x_t +W_{ho}h_{t-1}+W_{co}\circ  c_t+b_o)\\
h_t &= o_t \circ \tanh(c_t)\\
\end{aligned}
\label{LSTM_equation}
\end{equation}

\subsection{LSTM model for video prediction}
LSTM based recurrent neural network has been widely applied in the field of neural machine translation\cite{bahdanau2014neural,sutskever2014sequence,chung2016character}, video analysis\cite{xu2015show,johnson2015densecap,yeung2015end,yao2015describing}, etc. In these tasks formulated as a supervised learning problem, the goals are match a set of observation sequences to the correct target sequences or labels. However in many applications correspondences between videos or detailed labels are not available, exploring the spatial-temporal structure of the raw video sequences would be more appealing. 

 For $\mu$ECoG prediction, we used \cite{srivastava2015unsupervised} as the baseline model. The baseline model has a LSTM encoder, a LSTM decoder and a LSTM predictor. The encoder learns a compact representation for a certain number of observed frames, and the decoder reconstructs these observed frames from the encoded feature. The predictor then predicts future frames of the given sequence based on the encoder feature. The entire system can be learnt in an end-to-end manner based on the training sequences. To enhance the performance, instead of using one layer of LSTM for each submodule(encoder/decoder/predictor), multiple LSTMs are stacked to form more complex structures by adding nonlinearity. 
 
\begin{figure}
        \centering
                \includegraphics[width=0.5\textwidth]{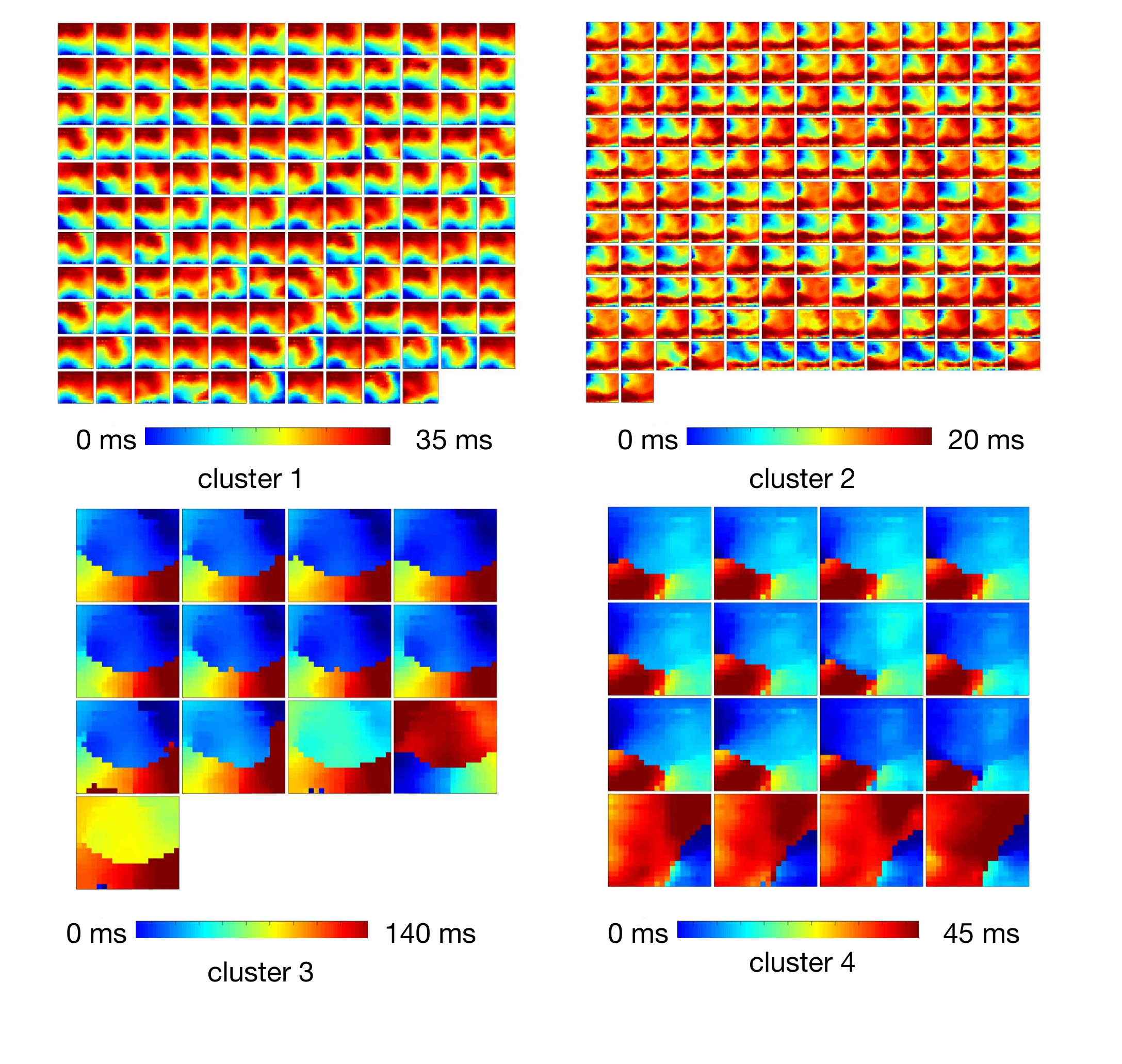}

\caption{Sample clusters of neural activity patterns. This figure shows the delay maps in several clusters identified using the method of \cite{song2015seizure}. The delay map captures how the apparent wave in a neural recording moves. Clusters 1-4 shown here correspond to upward, top-left to bottom-right,  counter-clockwise, and clockwise waves.}

\label{cluster neuron activity}
\end{figure}

Unlike the moving MNIST dataset \cite{srivastava2015unsupervised} used by \cite{finn2016unsupervised,kalchbrenner2016video,xingjian2015convolutional,srivastava2015unsupervised} for video sequence prediction, the $\mu$ECoG dataset have been observed to form multiple clusters, each with a distinct neural activity pattern, as shown in Fig.~ \ref{cluster neuron activity}. One way to exploit this multi-cluster nature of the $\mu$ECoG videos is by fitting a model for each cluster of sequences. This approach would require one to segment a long video into short sequences and furthermore, classify each sequence to one of predefined clusters. Such an approach is highly limited by the sequence segmentation and clustering. Besides, this pipelined framework is against the common approach of deep learning where one usually trains in an end-to-end manner.  Another alternative is by adding more LSTM cells. As the number of parameters grows in $O(n^2)$, $n$ being the number of LSTM cells in each layer, adding more LSTM cells is not efficient to fully exploit the clustered nature of the underlying signals. In the following subsection we propose a new approach to solve the assignment and optimization problem in an end-to-end manner.

\subsection{Multiple choice learning (MCL)} \label{section_MCL}
Typically, ensemble models are trained independently under different random initializations and prediction results are averaged during test time \cite{hinton2015distilling,lee2015m}. These models are commonly viewed as experts or specialists in the literature, although they are rarely trained to encourage diversity and specialization. In \cite{hinton2015distilling}, the authors distilled information from ensembles of convolutional neural network (CNN) by pre-clustering data. They pre-cluster the images in the dataset based on image categories, and each CNN specialist is only trained on a subset/cluster of images. During inference time, the prediction of label is by having a generalist model first determines the potential subcategory that input image might belong to and letting the ensemble models trained with this sub-category further determine the label of the image. Although this approach is sound for image classification where one typically has many labelled images, such an approach leveraging label-based pre-clustering is not feasible when facing unlabelled video dataset. Instead, we adopt the framework of Multiple Choice Learning (MCL) \cite{guzman2012multiple,lee2015m}, where the assignment of a training sample to each model is jointly solved with finding the optimal parameters for all models. 

In the video prediction setup, we have a set of models ${\theta_1, \theta_2, \cdots, \theta_M}$ such that $\theta_m : x_{(1:t-n)} \rightarrow \hat{x}_{(1:t)}$. $x_{(1:t-n)}$ are input frames in time 1 to $t-n$. $\hat{x}_{(1:t-n)}$ are the reconstructed frame from the input and $\hat{x}_{(t-n+1:t)}$ are the predicted frames in time $t-n+1$ to $t$. The loss for a sequence $x_{(1:t)}$ is defined in Eq.~\ref{ensemble_loss} , where $l(\theta (x_{(1:t-n)}),x_{(1:t)})$ is the mean square error loss between $\hat{x}_{(1:t)}$ and $x_{(1:t)}$. The goal of our MCL setup is to find the assignment variable $p_{im}$ and parameters for $\theta_m$ by solving the optimization problem defined in Eq.~\ref{ensemble_loss}. 
\begin{equation}
\begin{aligned}
&\displaystyle\min_{\theta_{(1:M)}, p_{im}} \displaystyle\sum_{i=1}^I \sum_{m=1}^M p_{im}l(x_i,\theta_m(x_i)) \\
&\text{s.t.} \displaystyle\sum_{m=1}^M p_{im}=1, p_{im} \in\{0,1\} \\
& \theta_m(x) : x_{1:t-n} \to \hat{x}_{1:t} \\
& l(x,\theta_m(x)) : ||x_{(1:t)} - \theta_m(x)||^2
\end{aligned}
\label{ensemble_loss}
\end{equation}
Note that in the training stage, at each iteration, we know the reconstruction and prediction accuracy of each current ensemble model on one instance $x_i$. Therefore, we can assign a training instance $x_i$ to the model that has the minimal reconstruction and prediction error. The optimization problem in Eq.~\ref{ensemble_loss} could be solved with a coordinate descent algorithm \cite{lee2015m} with stochastic gradient descent (SGD) shown below. The solution alternates between finding the assignment and optimizing the corresponding model's parameter. 

\begin{algorithm}
\caption{Coordinate Descent for MCL training of LSTM}
\begin{algorithmic}[1]
\State Dataset $D = \{ x_i\}$, SGD parameters $\lambda, \eta$
\State LSTM model parameters $\theta_1, \cdots, \theta_M$
\State Initialize $\theta_1,\cdots,\theta_M$ with pre-trained models
\State $ t \leftarrow 0$
\While{not converged}
\State $t \leftarrow t+1$
\State sample batch $ B \subset D$
\Procedure{Forward pass}{}
\State for $x_i \in B$, compute each model' loss: 
\State $l(x_i,\theta_m(x_i))$ defined in Eq.~\ref{ensemble_loss}
\State Update assignment variable $p_{im}$ as: 
\State $p_{im} = 1[[m= \displaystyle\text{argmin}_{m} l(x_i,\theta_m(x_i))]]$
\State $L_m = \sum_i p_{im} l(x_i, \theta_m(x_i))$
\EndProcedure
\Procedure{Backward pass}{}
\State for each $\theta_m$ apply gradient descent as :
\State $\theta_m \leftarrow \theta_m - \eta \bigtriangledown L_m -\lambda \bigtriangleup \theta_m$
\EndProcedure
\EndWhile\label{euclidendwhile}


\end{algorithmic}
\end{algorithm}

\section{Experiments}
We design experiments for $\mu$ECoG data prediction using multiple choice learning of an ensemble of LSTM. We first performed graph filtering \cite{shuman2013emerging} on $\mu$ECoG dataset to fill in the missing channels either caused by manufacturing defects or loss of contact on membrane. The graph transform basis for the $\mu$ECoG dataset is consistent across time so that the training set and testing set can share the same basis. This creates a spatially smoothed dataset and makes unsupervised LSTM prediction acurately. We compare the results obtained using the baseline single LSTM model, the randomly initialized LSTM ensemble model and MCL trained LSTM ensemble. We further improved the prediction accuracy by having another classifier choosing which model to use as predictor.

\subsection{Dataset}
We analyze $\mu$ECoG data
from an acute in vivo feline model of epilepsy. The 18 by 20 array of high-density active electrodes has 500$\mu$m spacing between nearby channels. The in vivo recording has a temporal sampling rate of 277.78 Hz and lasts 53 minute. We obtain a total of 894698 sequences each 20 frames long (10 for reconstruction and 10 for prediction, for visual display in the paper it is 20 for reconstruction and 20 for prediction) by applying a sliding window over the original video recording of 7 induced seizures. To get disjoint subset for training and testing, we choose one seizure period, and form the testing set by including all sequences from this seizure and the non-seizure period leading up to this seizure. We form the validation set by choosing another seizure period and including all sequences during that seizure period and the non-seizure period. All remaining sequences are included in the training set. In total we have 788627 training sequences, 64167 validation sequences and 41904 testing sequences. 

\subsection{Training the LSTM ensemble using MCL}

\begin{figure*}
        \centering
                \includegraphics[width=0.92\textwidth]{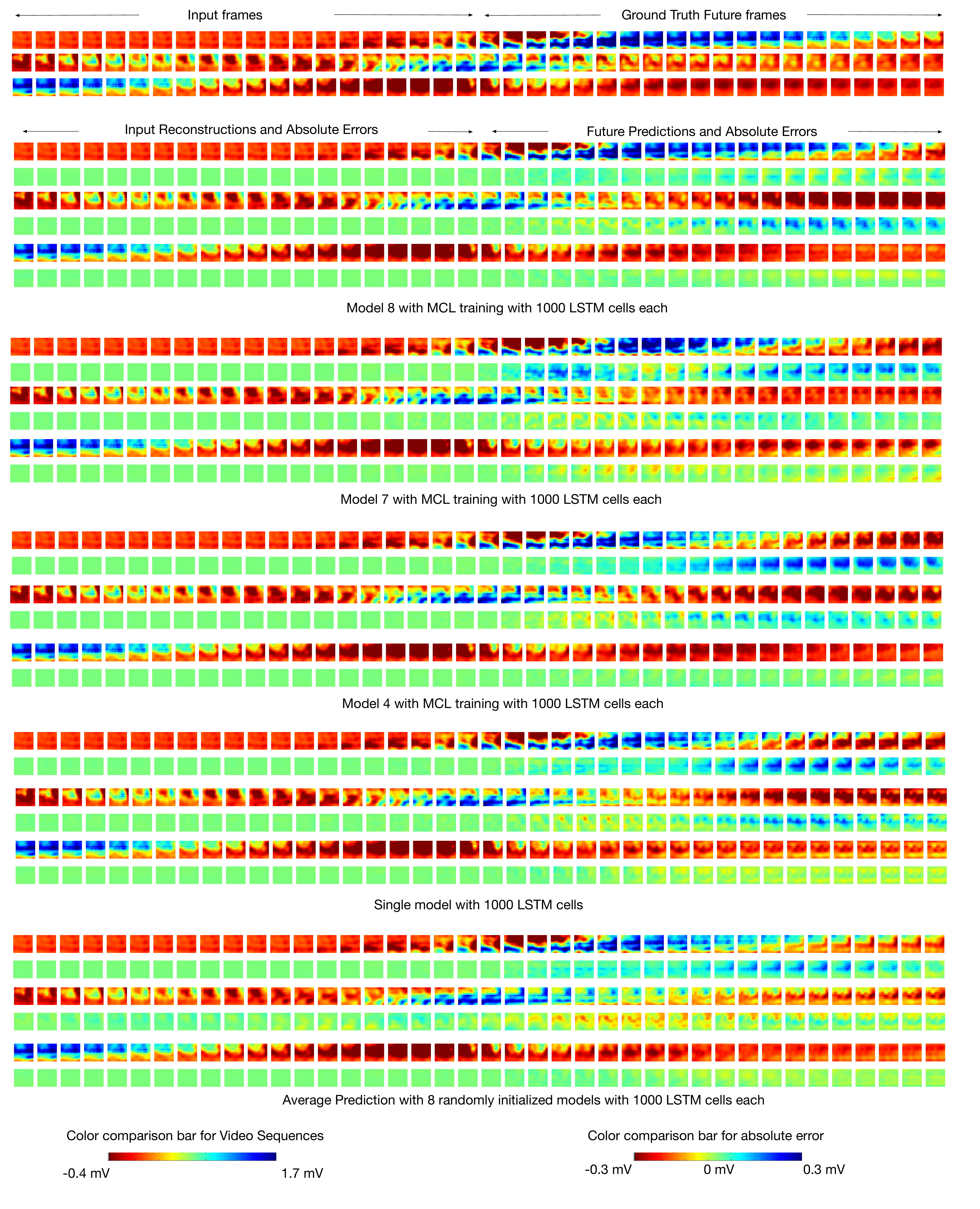}
                \caption{Reconstruction and prediction results for three test sequences by different methods. The top subfigure shows the original sequences. Each remaining subfigure contains the reconstructed frames and predicted frames for these three sequences by a particular model. Model 8, 7 and 4 are 3 models out of 8 models with MCL training that have lowest reconstruction error on these sequences respectively. The comparison against single model with 1000 LSTM cells and average prediction with 8 randomly initialized LSTM models are shown below.  The MCL training has led each model specializing at one kind of sequences by having lower prediction error. The absolute error plot against the ground truth demonstrates MCL training have lower prediction error.}
                               \label{sequences results}
\end{figure*}

We train a LSTM ensemble with 8 models. For parameter initialization, we first try random initialization for all 8 models. But we find once the gradient descent is made for the first mini-batch, one model is much better updated comparing to the rest and this model would have the lowest error for the majority of the remaining mini-batches. This causes only one model gets updated during training most of time. To overcome such problem, we randomly divide the training set into 8 non-overlapping subsets. Initialize one model with one of subset. We train all models using each subset by minimizing the mean square error loss using back propagation through time and SGD with a learning rate of $2*10^{-3}$ and momentum of 0.9. Dropout is applied only on non-recurrent connection as suggested \cite{zaremba2014recurrent}. We only train one epoch for each to ensure sufficient diversities between models. We then train all 8 models jointly using the MCL method described in Section \ref{section_MCL} and perform early stopping base on error of the validation set.

Each LSTM model has the same structure as \cite{srivastava2015unsupervised}, with two LSTM layers each with 1000 nodes. For MCL training, we use 4 Nvidia k80 GPUs in a cluster for training. Since the loss function is coupled with all models and could not be trained in a sequential manner. To enable our experiment scale, we use Message Passing Interface (MPI) standard to enable high speed GPU communication. Each GPU loads two models. As a comparison to MCL training, we also train three benchmark models.  The first benchmark model consists of two LSTM layers, each with 1000 nodes. The second benchmark model has 3000 nodes each layer. The second benchmark model has roughly similar amount of parameters as the ensemble with 8 LSTM models. We also train another benchmark of 8 random initialized 1000 nodes LSTMs and use the average of the prediction results by all 8 models as the final predicted signal.

\begin{figure}
        \centering
                \includegraphics[trim={4cm 0.5cm 1cm 1cm},clip,width=0.5\textwidth]{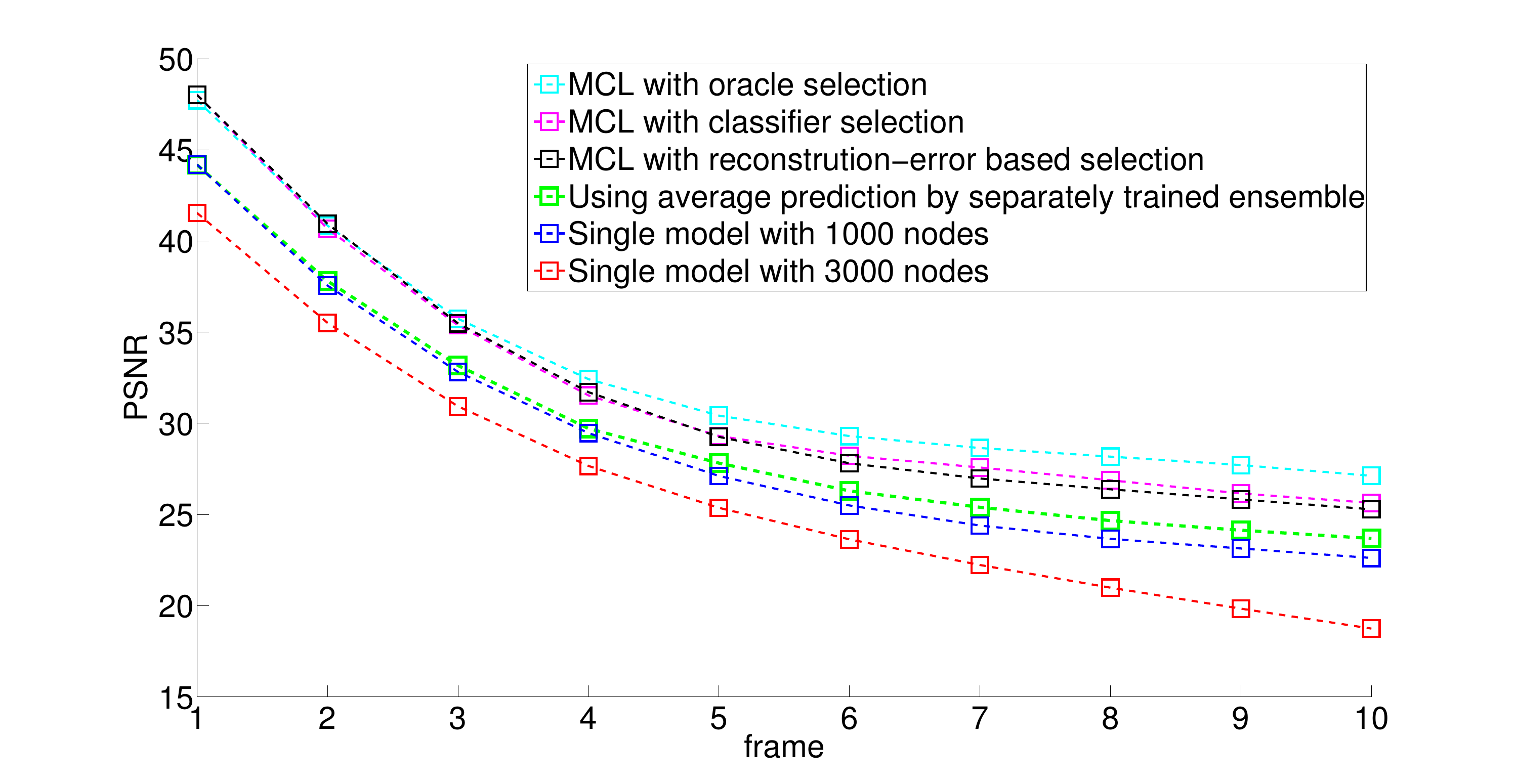}
                \caption{Peak signal noise ratio (PNSR) against prediction time with different methods of test set.}
                \label{PATC}
\end{figure}

Sample prediction sequences of testing datasets are shown in Fig.~\ref{sequences results}.  Model 8, 7 and 4 are models that have the lowest reconstruction errors on those sequences respectively, and the best model in terms of prediction accuracy also have the lowest reconstruction error in the case shown in here. This shows the model diversity trained with MCL.
The prediction accuracy against time comparison is shown in Fig.~\ref{PATC}. The PSNR is defined as:
\begin{equation*}
PSNR = 10*log_{10}(\frac{(max_I^2)}{MSE})
\label{PSNR_def}
\end{equation*}
Where MSE is the mean square error of prediction frames against ground truth frames and $max_I$ is the maximum intensity of the dataset. The oracle selection shown in Fig.~\ref{PATC} uses the model that has the lowest prediction error.  Since ground truth future frames are not available during inference, such selection mechanism is not practical in reality. The reconstruction-error based model selection chooses the model that has the lowest reconstruction error. The short term prediction accuracy between oracle selection and reconstruction-error based selection are roughly the same, but the accuracy of the latter drops faster than oracle selection as the prediction horizon increases. Even so the reconstruction-error based selection still beats the closest benchmark of average prediction with randomly initialized ensemble by a large margin.

From Table~\ref{prediction statistics datasets}, it is clear that the 3000 nodes LSTM model is worse than other benchmarks. Because the model does not have any structure to exploit the multi-cluster nature of neural activities, simply adding more nodes makes the number of parameters to be trained grow in an exponential manner. It is less likely to converge to a good local minimum as such model is prone to overfit the training set.

\subsection{Model selection as classification}
To further enhance from reconstruction-based selection, we train a multilayer perceptron (MLP) classifier to select which model to use for prediction. The classifier takes the concatenated LSTM hidden features at the last input frame from all models as input and output the probability of the best LSTM model to use as predictor. The input to MLP classifier is 8000 (1000 dimension feature per model). We use batch normalization\cite{ioffe2015batch} as regularization and used  three fully connected layers. By using the model that is predicted to have the highest probability by the classifier, we obtain a slight improvement compared to using the reconstruction-error based selection shown in Table \ref{prediction statistics datasets}.

 \begin{table}
  \centering

   \scalebox{0.85}{
    \begin{tabular}{  |p{7cm} |  p{2 cm}|}
    \toprule 
    model & PSNR \\
    \toprule
    MCL with oracle selection  & 32.2626 \\
    MCL with classifier selection & 31.2767\\
    MCL with reconstruction-error based selection  & 31.0636 \\
    Using average of prediction by separately initialized ensemble & 29.0722 \\
    Single model with 1000 nodes & 28.3495\\
    Single model with 3000 nodes & 25.8128 \\
    \bottomrule
    \end{tabular}
    }
          \caption{PSNR over all predicted frames with different methods for test set. The MCL training and random initialization both have 8 models. Each ensemble models have 1000 nodes.}
    \label{prediction statistics datasets} 
\end{table}

\subsection{Relationship between trained models with neural activity patterns}

\begin{figure}
        \centering
                \includegraphics[trim={2cm 0.5cm 1cm 1cm},clip,width=0.5\textwidth]{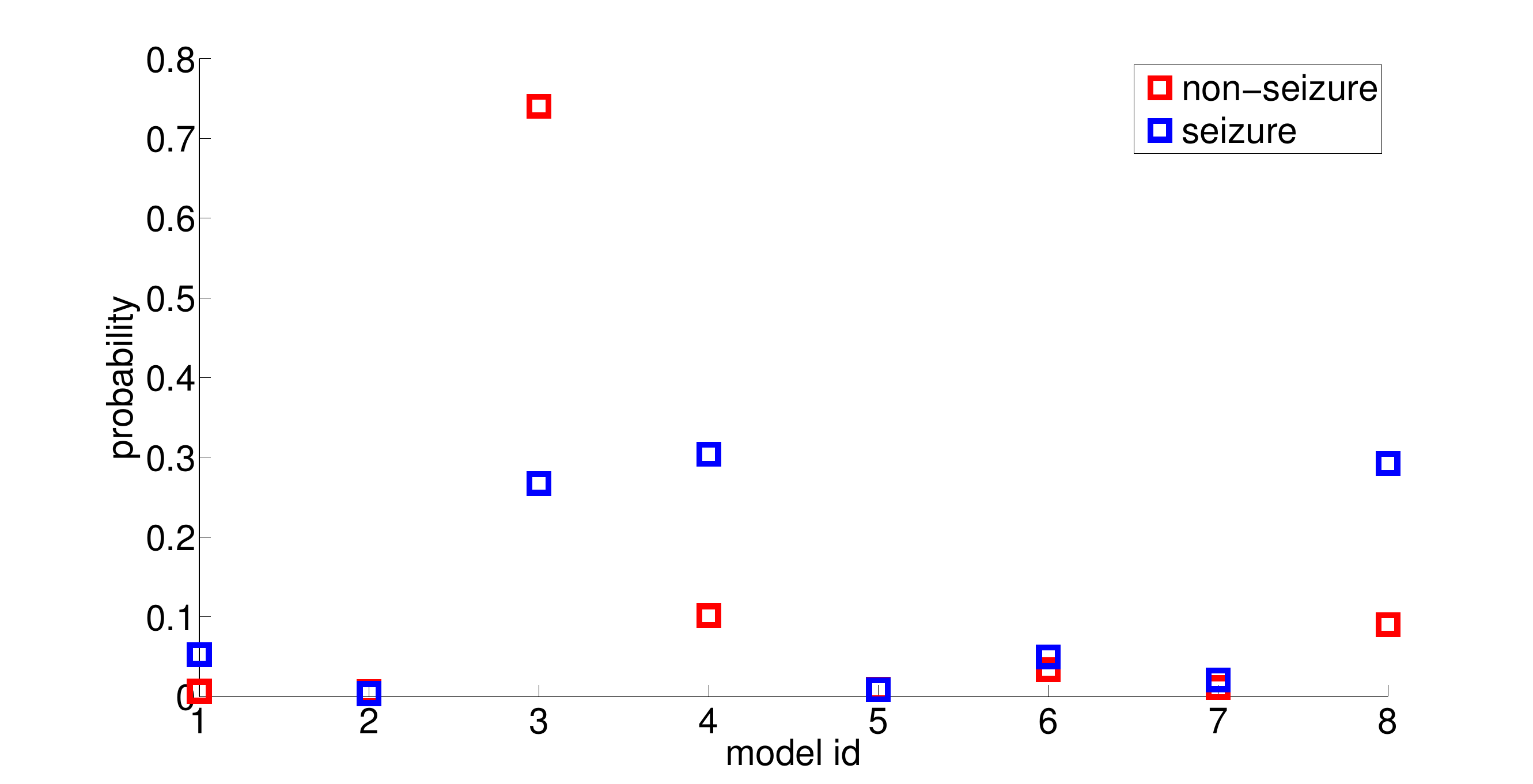}
                \caption{Probability of model selected for video prediction in non-seizure/seizure stage of test set.}
                \label{prob_model}
\end{figure}

\begin{figure*}
        \centering
                \includegraphics[width=0.98\textwidth]{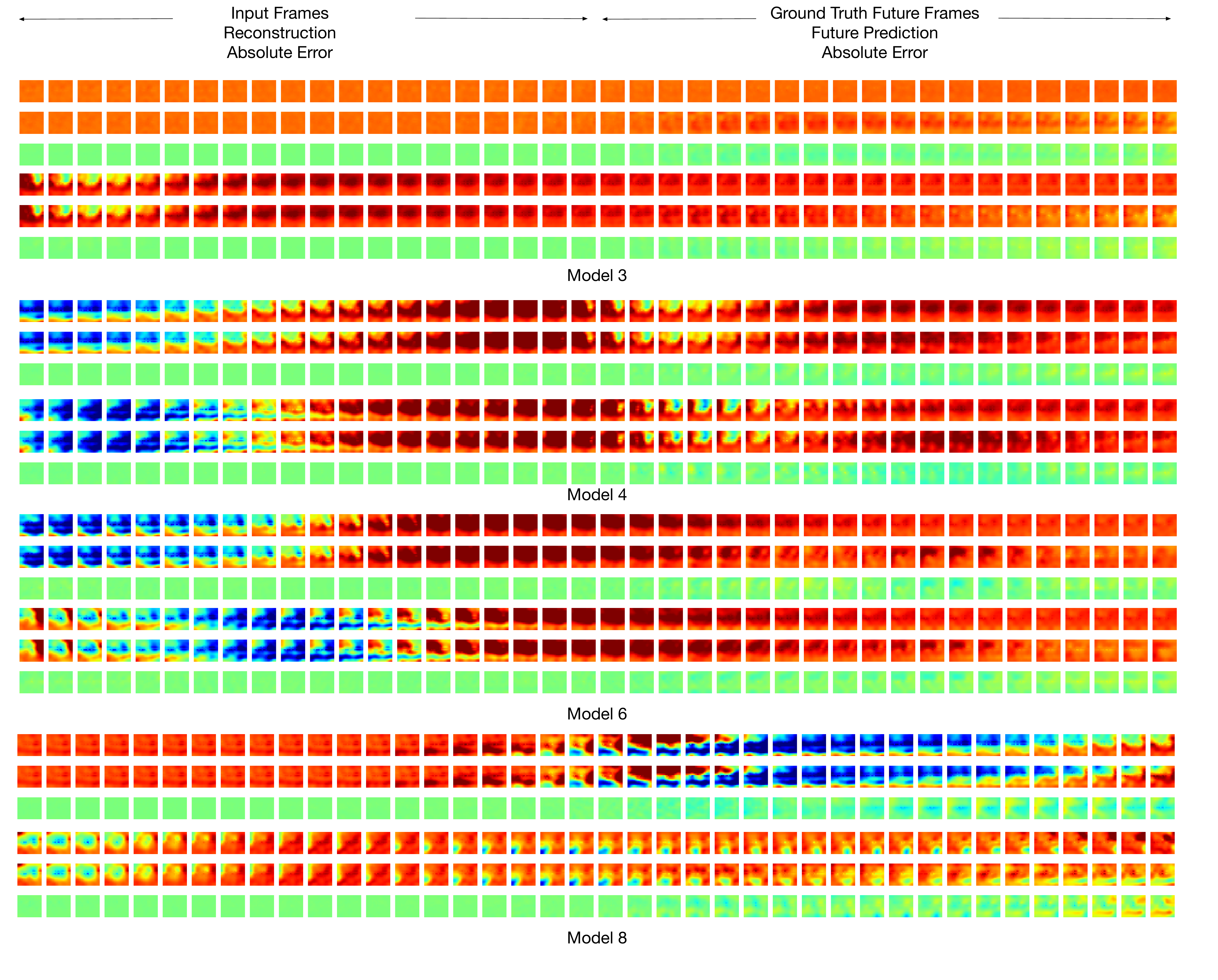}
                \caption{4 models prediction specialties in the ensemble trained with MCL. Each example sequence is shown with the vertical order of ground truth, reconstruction/prediction and absolute error. }
                \label{cluster results}
\end{figure*}

In this section, we analyze the potential relationship between different models in the learned ensemble and different neural activity patterns during seizure and non-seizure durations. For each testing sequence, we assigned the model based the oracle selection (i.e. the model with the least prediction error). Fig.~\ref{prob_model} shows the probability of different models. The difference between seizure and non-seizure stage shows there are essentially different neural activities in these stages.
We see that most neural  activity patterns during non-seizure periods can be captured by model 3, whereas there are several different  clusters of activity patterns during seizure periods, mainly captured by models 3, 4, 6 and 8. We further investigate the types of neuron activities captured by these models. We find that model 3 is good at predicting silent neural activity namely most of neurons will be at resting potential (hence this model is used in both non-seizure and seizure periods). Model 4 is good at predicting neural activities restricted to a small region. Model 6 is good at predicting when most neurons are going into refractory period after action potential. And model 8 is good at predicting moving neural activity patterns. Such patterns are more common in seizure stage, which explains why model 8 are selected more often during seizure stage. Those patterns are shown in Fig.~\ref{cluster results}.

Table~\ref{non_seizure_trans} shows the transition probability of models between consecutive time windows for non-seizure stage and seizure stage in test set. The transition probability is defined as: \\
$ P( \underset{m}{\operatorname{argmin}}(l(x_i,\theta_m(x_i))) |  \underset{m}{\operatorname{argmin}}(l(x_{i-1},\theta_m(x_{i-1}))))$
Where $x_{i-1}$ and $x_{i}$ are sequences corresponding to adjacent sliding windows. The diagonal elements of the transition matrix shows the likelihood that the same model is selected for predicting the next sequence. The high self-transition probability shows each model in the ensemble has quite stable prediction power within a short period. As neural activities get more complex from non-seizure to seizure, the transition between models are more frequent demonstrated by the reduction of self-transition probabilities from non-seizure to seizure stage. The high transition probability between models 4 and 8 in both stages indicates the global wave propagation is highly likely to be followed by another local active potentials (see sequences of model 4 in Fig.\ref{cluster results}) and vice versa. And high transition probability from model 6 to model 3 demonstrate the transition of neuron from refractory period into resting state. 

\begin{table}

     \scalebox{0.7}{
   \begin{subtable}{.5\textwidth}
   \centering
  
    \begin{tabular}{  |p{1cm} |  p{1 cm}|p{1cm} |  p{1 cm} |p{1cm} |  p{1 cm} |p{1cm} |  p{1 cm} |}
    \toprule 
    model 1& model 2&model 3&model 4&model 5&model 6&model 7&model 8 \\
    \toprule
    0.4819  &   0.0433 &   0.0015 &   0.0106  &  0.0363 &   0.0058  &  0.0066 &   0.0046 \\
    0.0543 &   0.4882  &  0.0005  &  0.0097  &  0.0259  &  0.0108  &  0.0088  &  0.0081 \\
    0.1014  &  0.1339  &  0.9716  &  0.0595  &  0.0751  &  0.1974 &   0.1140  &  0.0535 \\
    0.1413  &  0.0984  &  0.0103  &  0.7039 &   0.1192  &  0.0238  &  0.1776  &  0.1872 \\
    0.0507  &  0.0315  &  0.0013  &  0.0128 &   0.6218  &  0.0079  &  0.0088 &   0.0049 \\ 
    0.0326 &   0.0591  &  0.0074  &  0.0172 &   0.0259  &  0.7378 &   0.0110  &  0.0081 \\ 
    0.0471  &  0.0472 &   0.0024  &  0.0133 &   0.0259  &  0.0043 &   0.4539 &   0.0220 \\ 
    0.0906  &  0.0984 &   0.0050  &  0.1731  &  0.0699  &  0.0122 &   0.2193 &   0.7115 \\    
    \bottomrule
    \end{tabular}   
    \captionsetup{justification=centering}
     \caption{non-seizure}
    \end{subtable}%
   }
   \\
   
\scalebox{0.7}{
     \begin{subtable}{.5\textwidth}
     \centering
  
    \begin{tabular}{ | p{1cm} |  p{1 cm}|p{1cm} |  p{1 cm} |p{1cm} |  p{1 cm} |p{1cm} |  p{1 cm}| }
    \toprule 
    model 1& model 2&model 3&model 4&model 5&model 6&model 7&model 8 \\
    \toprule
 0.4286  &       0  &  0.0573  &  0.0126    &     0 &   0.0769  &  0.0909  &  0.0164 \\
         0  &  0.2500     &    0  &  0.0063  &       0    &     0   &      0  &  0.0033 \\
    0.0893   & 0.2500  &  0.7993  &  0.0252  &  0.1111  &  0.3077   & 0.3636 &   0.0559 \\
    0.0893  &  0.2500  &  0.0681 &   0.7413  &  0.1111 &   0.0385  &  0.1364  &  0.1678 \\
         0     &    0   & 0.0072  &  0.0063  &  0.4444 &   0.0192   &      0    &     0 \\
    0.1071   &      0    &     0   & 0.0536  &  0.1111  &  0.5385   &      0     &    0 \\
    0.0179   &      0  &  0.0072  &  0.0063 &   0.1111   &      0  &  0.2727  &  0.0329 \\
    0.2679 &   0.2500  &  0.0609  &  0.1483 &   0.1111  &  0.0192 &    0.1364 &   0.7237 \\
    
    \bottomrule
    \end{tabular}
    \captionsetup{justification=centering}
     \caption{seizure}
    \end{subtable}
    }  
        \caption{Oracle model selection transition matrix in non-seizure and seizure stages in testing set. Each column shows the probability of the next model being selected for video prediction for adjacent observation window.}

    \label{non_seizure_trans}    
\end{table}

\section{Conclusion}
In this work, we have successfully applied the deep learning approach to the challenging problem of predicting neural activities observed by high resolution $\mu$ECoG.
We formulate the problem as a video prediction problem. Observing that there are multiple clusters of neural activities, we propose an extension of MCL from CNN to LSTM models. The MCL solves the assignment problem jointly with the loss minimization problem. The MCL has enabled a significant improvement in video prediction accuracy compared to averaging the predictions by separately trained LSTM models. Some of the models indeed are found to be able to model different motion patterns in the neural dataset. We find that using the reconstruction error as a guideline to select the model to be used for prediction can yield predictions close to that  using an oracle selected model. Using a trained classifier for model selection further improves prediction accuracy slightly. Finally, we conduct an analysis of the association between the models selected and the neural activities of the underlying video sequences. The analysis reveals the differences in the distribution of selected models and the model transition probability matrix between seizure and non-seizure stages.

\section{ACKNOWLEDGEMENT}
This work was funded by National Science Foundation award CCF-1422914.
 \vspace{-0.05in}

{\small
\bibliographystyle{ieee}
\bibliography{Diversity_LSTM}
}

\end{document}